# Development of Anthropomorphic Arm for collaborative and Home Service Robot CHARMIE

HAFIZ SYED FAWAD ADIL[1], TIAGO RIBEIRO [1], ANTONIO RIBEIRO [1], CAROLINA LOPES [1], JOANA COELHO [1], GIL LOPES[2], FERNANDO RIBEIRO[1]

[1] *Laboratório de Automação e Robótica, University of Minho, Campus de Azurém, 4800-058 Guimarães, Portugal*

*fawadsyedadil@gmail.com, id9402@alunos.uminho.pt, id11906@alunos.uminho.pt, pg50176@alunos.uminho.pt, id8667@alunos.uminho.pt, fernando@dei.uminho.pt,*

[2] *GECAD - ISEP*

*Technical University of Porto*

*Porto, Portugal*

*agl@isep.ipp.pt*

**Abstract**

Service Robots are becoming a reality and is progressing by leaps and bound. Alongside this avenue of robotics is also the field of Prosthesis which is evolving rapidly. These topics are hot cake in the industry and are widely developed. Some having one quality and some having others but making a shoreline towards success. The development of arm and hand as individual or as a package is becoming common as more and more advanced collaborative robots come to surface and join their human counter parts in the real world. This dissertation presents an open source 3D printed arm which was assembled and programmed on two fronts. The first is the control of the hand by utilizing tele operation by using a camera and an ML based hand pose estimation. Secondly, is the control via kinematics utilizing Denavit-Hartenberg method to define the motion and to find the position of the end effector in 3D space.

## 1. INTRODUCTION

In recent years, Robots are becoming more and more involved in daily human lives. They are crucial for human interaction and tasks. With increase in demand of robots that can work alongside human in domestic and industrial application, a lot of research at many labs are being done and producing wondrous results. To operate in an anthropomorphic fashion, a humanoid robot needs to have enough DOF (Degree of Freedom) to execute most human actions. For arms, 7-DoF from the shoulder to the wrist allows the performance of most human motions, and for hands, this number varies from 5 to 24 under actuated joints, depending on the type of motion required [1]. A lot of examples can be seen in use. NASA's Robonaut Arm and ARMAR-III [2] humanoid platforms are very good examples of such robots. For humanoid arms to perform human like motion, at least 7 DOF are required and for hands a minimum of 5 DOF are required to define the motion. A very good example for the arm is the LIMS2 [3] which introduces a lightweight robot arm beyond the shoulder. JUSTIN [4] is another very good example for such robots.

Since this work is done under the development of CHARMIE [5]. CHARMIE is targeted towards domestic robot as home service and assistant robot. The social goals for this work are also to use this robot in places like nursing homes where it can interact with people and perform tasks. With respect to this project, the arms were needed to be selected and control strategies were needed to be developed. Inmoov [6] is an open-source robot with huge internet following and community. This arm is a 10 DOF robotic arm consisting of 5 DOF hand, 1 DOF wrist, 1 DOF elbow and 3 DOF shoulder.

The work is divided in to three parts. Firstly, the assembly and the avenues where cost can be reduced and changes that will improve the performance of the arm. Secondly define the kinematics of the arm and find the limits of arm operation and to design control software to actuate the arm and verify the motion. Thirdly, to design a teleoperation method to control the hand of the arm using computer vision and ML (Machine Learning) based hand pose estimation.

# 2. METHODOLOGY

The methodology discusses the three aspects of the work in individual aspects and what work has been done to achieve the results

## 2.1. Electromechanical Assembly and Modification

Since Inmoov is an open-source robot, the work started by downloading the files for the left arm and hand and all components are 3D printed as per the inventor's specification. PLA (Polylactic Acid) is used for printing material on Makerbot Replicator 2. Some components were modified and reprinted as the necessity came. The first modification that is done is to replace the part servo-pulley (part names as per Inmoov). Original pulley is a single-track pulley and as per assembly instructions, the tension in the tendons is not strictly taut which leads to loose tendons and imprecise movements. To keep the finger's tendons taut, the single-track pulley is replaced by a dual track pulley. The tendon motors are rotated in either way and the tendons are tightened to keep the fingers in a taut position and avoid loose movement. The second modification is the GearHolder and gearpotentio. These two components are needed to keep the elbow potentiometer in place. The issue was the size of the components and was resolved reprinting with 100% increase in size. The second modification is the replacement of the motors i.e., servo motor of elbow and arm rotate. Original design used Hitec HS-805BB servo, and these were replaced by TD-8125MG. The new motor has similar torque output but in a smaller form factor. These changes needed a modification in the mounting methods. Two mounting plates are designed in Autodesk Inventor. A new part called bracket is designed to mount the arm to a platform. Since this part is based off the original parts, it houses the arm, the arm lift piston, and the potentiometer for the arm lift position. FEA (Finite Element Analysis) are performed, and parts were found satisfactory with safety factor of 15. The electronics assembly consist of three small circuits comprising of header pins for servos and fuse for safety. Arduino Mega is used as the controller because it has a lot of PWM pins that are needed for servo actuation. There are a total of 10 servo motors, 6 motors are MG996R which control the actuation of fingers and wrist, 2 motors are TD-8125MG which control the actuation of elbow and the arm rotate motion and 2 motors are Hitec HS-805BB for shoulder lift and shoulder rotate joint. These motors are actuated individually to find the extreme angles. The angles are measured under two criteria. Firstly, the current consumed by the motor and the position of the robot. The robot position is important to avoid collisions. These angles are used in the Kinematic calculation, Computer vision calculations and actuation of the arm. The total mass of 3D print came out to be 1864 grams. The mass of the motors altogether is 804 grams. The total mass of the arm is about 2700 grams.

The table consisting of the angle values are as below

TABLE 1. Joint rotation angle range

| | Joint Name | Minimum Angle Value | Maximum Angle Value |
|---|---|---|---|
| 1 | Shoulder Lift | 6 | 55 |
| 2 | Arm Rotate | 30 | 160 |
| 3 | Bicep Rotate | 70 | 120 |
| 4 | Elbow | 135 | 45 |
| 5 | Wrist | 30 | 150 |

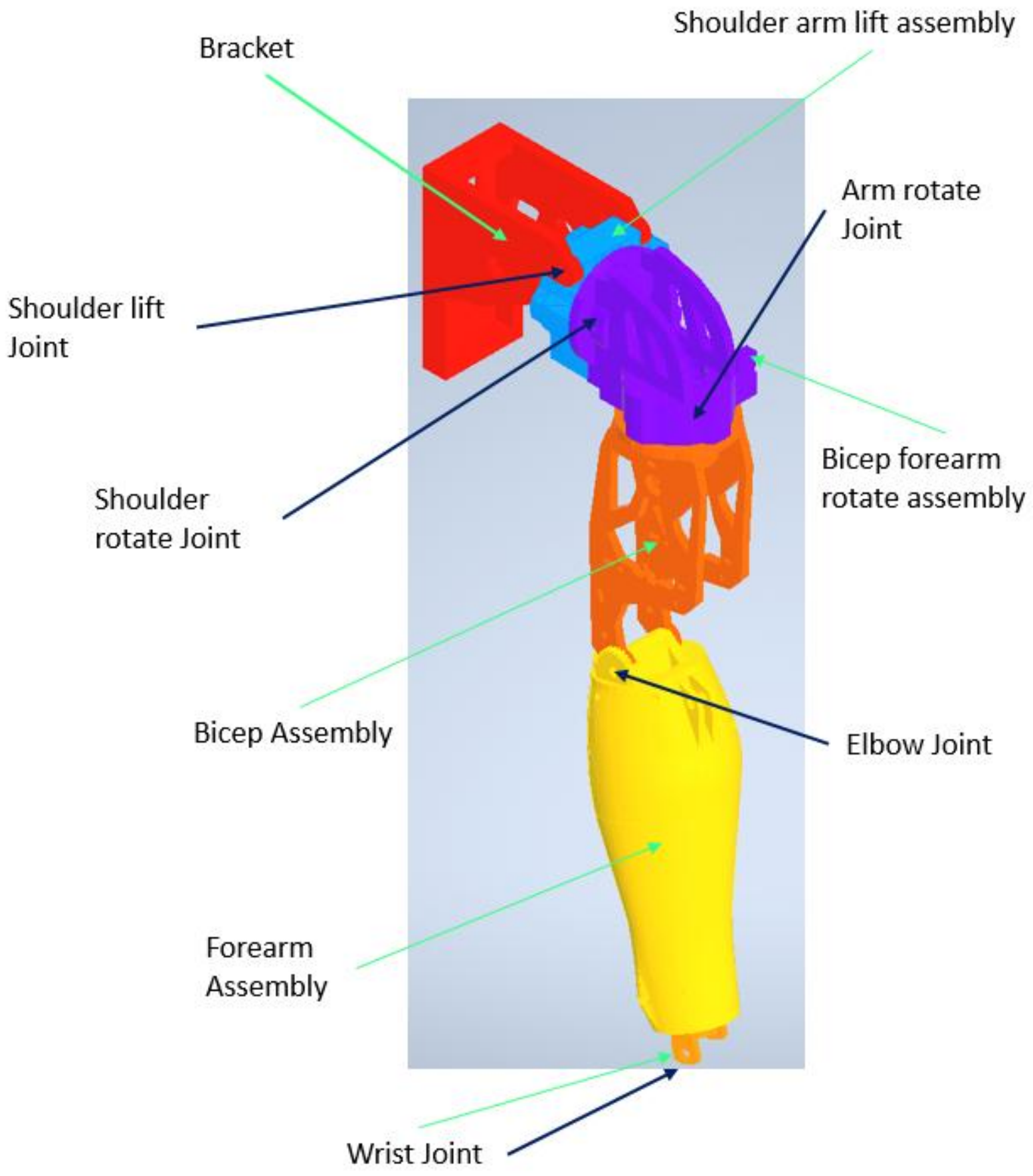

Fig. 1 Arm components in CAD environment

The servo angle for the fingers of the hand are as follows

TABLE 2. Finger Servo Angle

| | Finger name | Maximum Angle Value | Minimum Angle Value |
|---|---|---|---|
| 1 | Thumb | 170 | 70 |
| 2 | Index | 25 | 125 |
| 3 | Middle | 30 | 135 |
| 4 | Ring | 30 | 120 |
| 5 | Small | 30 | 110 |

### 2.2. Technical Specifications

The kinematics developed are for arm only. First step in this work is the creation of assembly in a CAD system for finding the centers of rotation and the dimensions of the joints. The arm in CAD is shown in figure 1 with the all the joint and part names.

The arm is a 5 DOF arm and is calculated by Grubler's Rule

$$DOF = \lambda(N - J - 1) + \sum_{i=1}^{J} F_i \quad (1)$$

Where N is the number of links, J is the number of joints, Fi is the DOF at the ith joint and λ is a constant which is either 6 for 3D motion or 3 for 2D motion. The DOF for the robot is calculated below as in (1)

DOF=6(6-5-1)+5

DOF=5

The DH equations as below show the transformation matrix for each link and the final transformation matrix is calculated as below

$$T_i^{i-1} = R(Z_{i-1}\theta_i)T(Z_{i-1}d_i)T(x_i\alpha_i)R(x_i\alpha_i) \quad (2)$$

$$=$$

$$\begin{bmatrix} C\theta_i & -S\theta_i & 0 & 0 \\ S\theta_i & C\theta_i & 0 & 0 \\ 0 & 0 & 1 & 0 \\ 0 & 0 & 0 & 1 \end{bmatrix} \begin{bmatrix} 1 & 0 & 0 & a_i \\ 0 & 1 & 0 & 0 \\ 0 & 0 & 1 & d_i \\ 0 & 0 & 0 & 1 \end{bmatrix} \begin{bmatrix} 1 & 0 & 0 & 0 \\ 0 & C\alpha_i & -S\alpha_i & 0 \\ 0 & S\alpha_i & C\alpha_i & 0 \\ 0 & 0 & 0 & 1 \end{bmatrix}$$

$$=$$

$$\begin{bmatrix} C\theta_i & C\alpha_i S\theta_i & S\alpha_i S\theta_i & a_i C\theta_i \\ S\theta_i & C\alpha_i C\theta_i & -S\alpha_i C\theta_i & a_i S\theta_i \\ 0 & S\alpha_i & C\alpha_i & d_i \\ 0 & 0 & 0 & 1 \end{bmatrix}$$

Where C and S are SINE and COSINE, θ is the angle of rotation around Z-axis, α is the angle offset around X-axis of the joint, a and d are joint offset is X and Z axis respectively and i represents the link number in (2)

Each of the individual transformation matrix is then multiplied as shown below

$$H = T_i^0 = T_1^0 T_2^1 \dots T_t^{i-1}$$

$$=$$

$$T_i^0 = \begin{bmatrix} R_i^0 & T_i^0 \\ 0 & 1 \end{bmatrix} \quad (3)$$

Where T is transformation matrix, i is the link number. H represents the total transformation consisting of 3x3 rotation matrix represented by R to represent orientation of the end effector and a 3x1 Translation matrix represented by T from the base joint as in (3).

Utilizing the CAD assembly and following the Denavit – Hartenberg Method (DH) [7] were realized for the arm as in the table below

TABLE 3. DH Parameters for the arm

| Link i | θ(°) | d(mm) | a(mm) | α(°) |
|---|---|---|---|---|
| 1 | $\theta_1$ | 0 | 11.783 | -90 |
| 2 | $\theta_2$-90 | 66.104 | 0 | -90 |
| 3 | $\theta_3$-90 | 224.60 | 0 | -90 |
| 4 | $\theta_4$+26.6 | 0 | 0 | 90 |
| 5 | $\theta_5$ | 289.5 + 80.5 = 370 | 0 | 0 |

The d parameter for link 5 has two values added together, 289.5 is the distance from the elbow to the wrist joint but since the end effector i.e., the hand is also included so 80.5 is added to put the end effector at the center of the hand.

These parameters in table were realized by using the kinematic axis diagram in figures below.

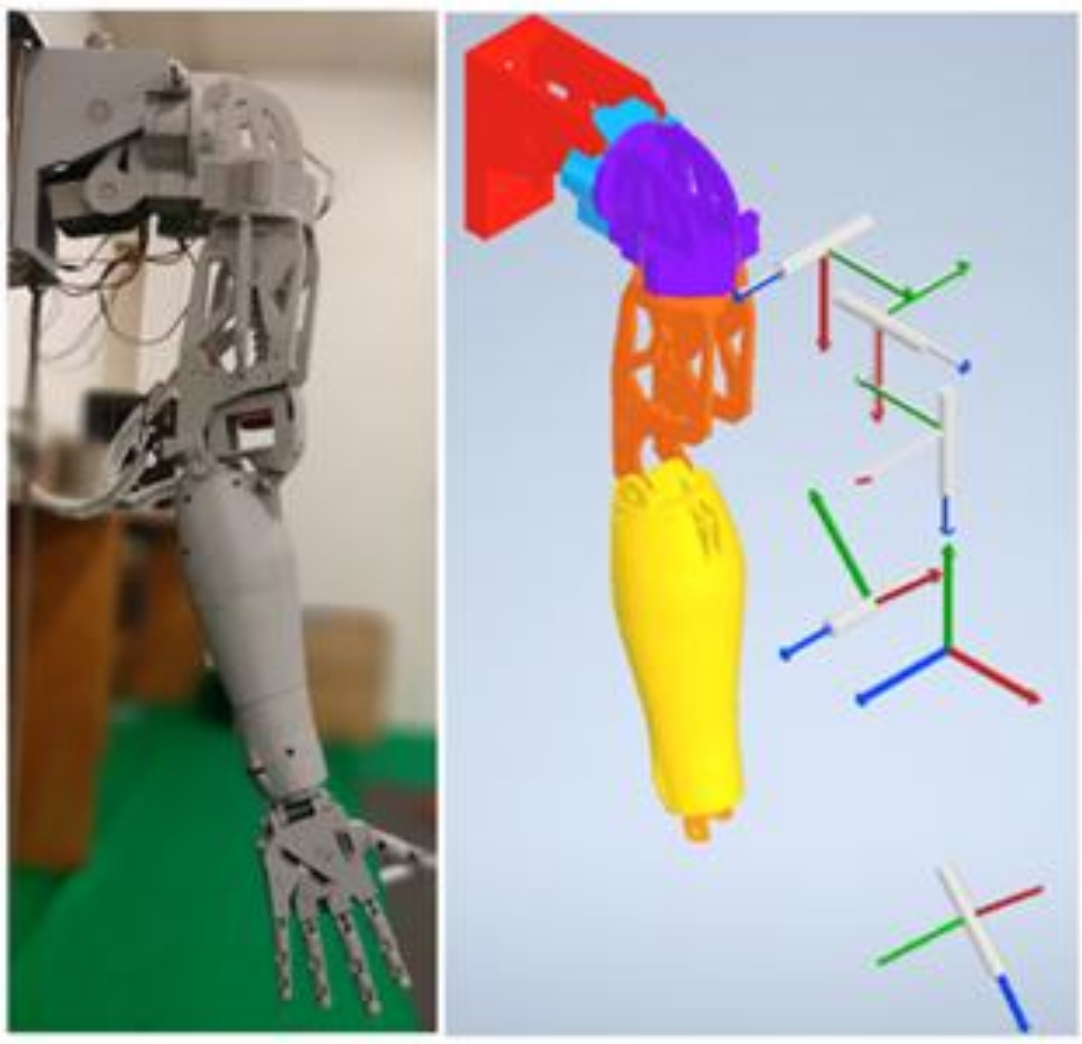

Fig. 2 Arm Kinematics and the joint axes

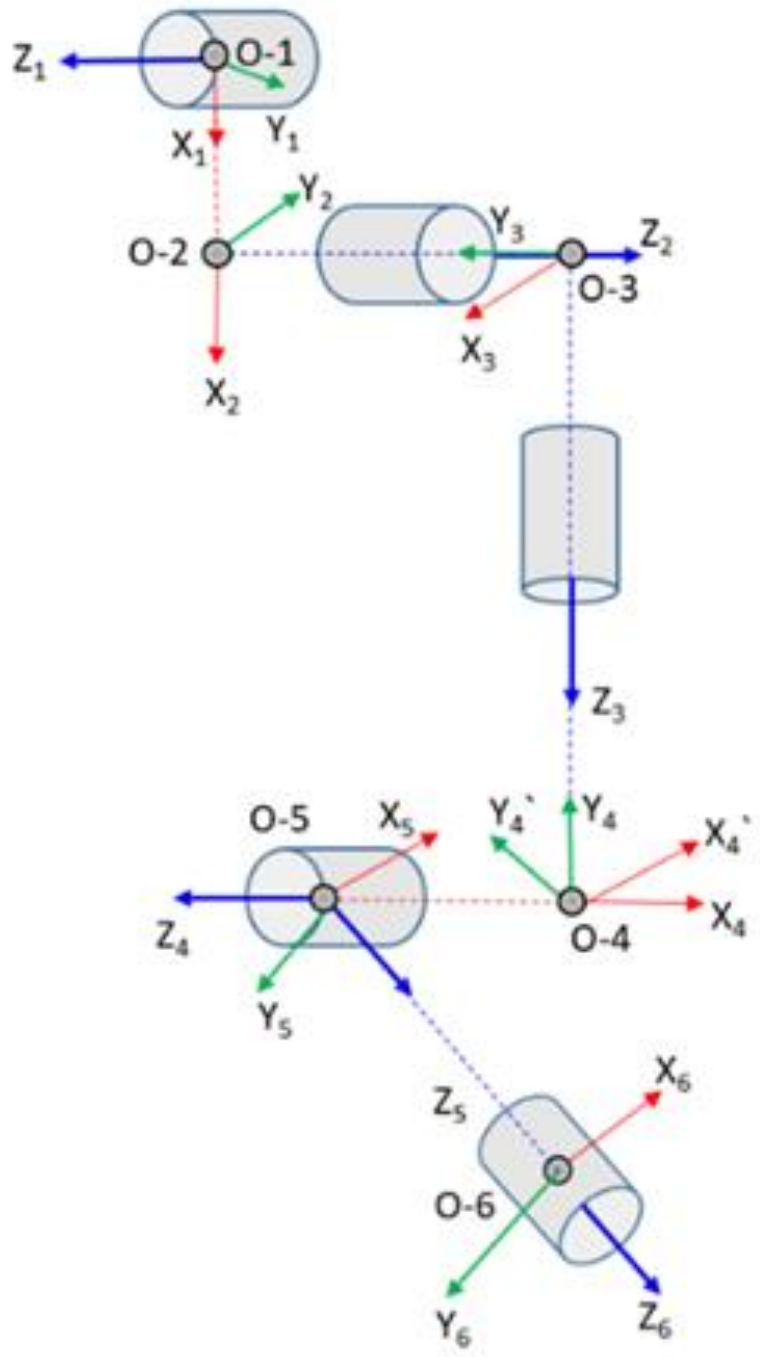

Fig. 3 Arm joint axes for DH Parameters

The parameters in Table 1. are realized using an opensource software called MRPT (Mobile Robot Programming Toolkit) [8]. It helped finalizing the parameters and was a verification tool for the software developed in Python to control the robot. The MRPT model is as below

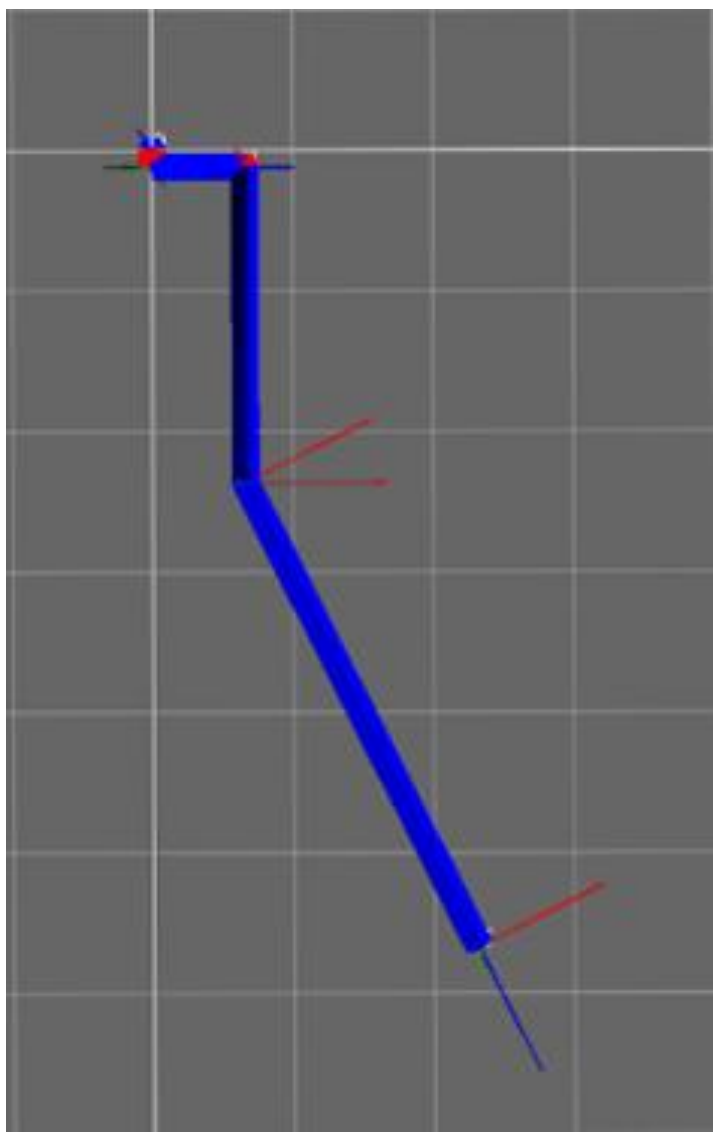

Fig. 4 Arm simulation in MRPT

### 2.3. Mimicking through Vision

The main software used for this part of the work is OpenCV which can be used with Python. The initial approach was to develop a software to detect hand in a frame and detect the pose. After studying a bit more, the idea of making a software from root up was dropped and Google Mediapipe [9] was adapted. A few perquisite software were needed to enable the robot to mimic human hand movement. As the robot is actuated by using Arduino as the controller for servo motors, PyFirmata [10] for creating serial communication. The software detects the Arduino's port and setup the servo motors as output. Google Mediapipe needs two values for setting up the detection and tracking confidence. For this work 0.7 was selected out of 1. Several values were tested with, for a value of 0.5, the detected hand pose shifted randomly and any value above 0.7 was very restrictive and stopped detecting hand in some lighting condition. 0.7 was also selected as it gives a decent frames per second of 16 – 25 fps. The program is fed the frames form the camera and a rudimentary bounding box was created to keep the hand or hands in perspective.

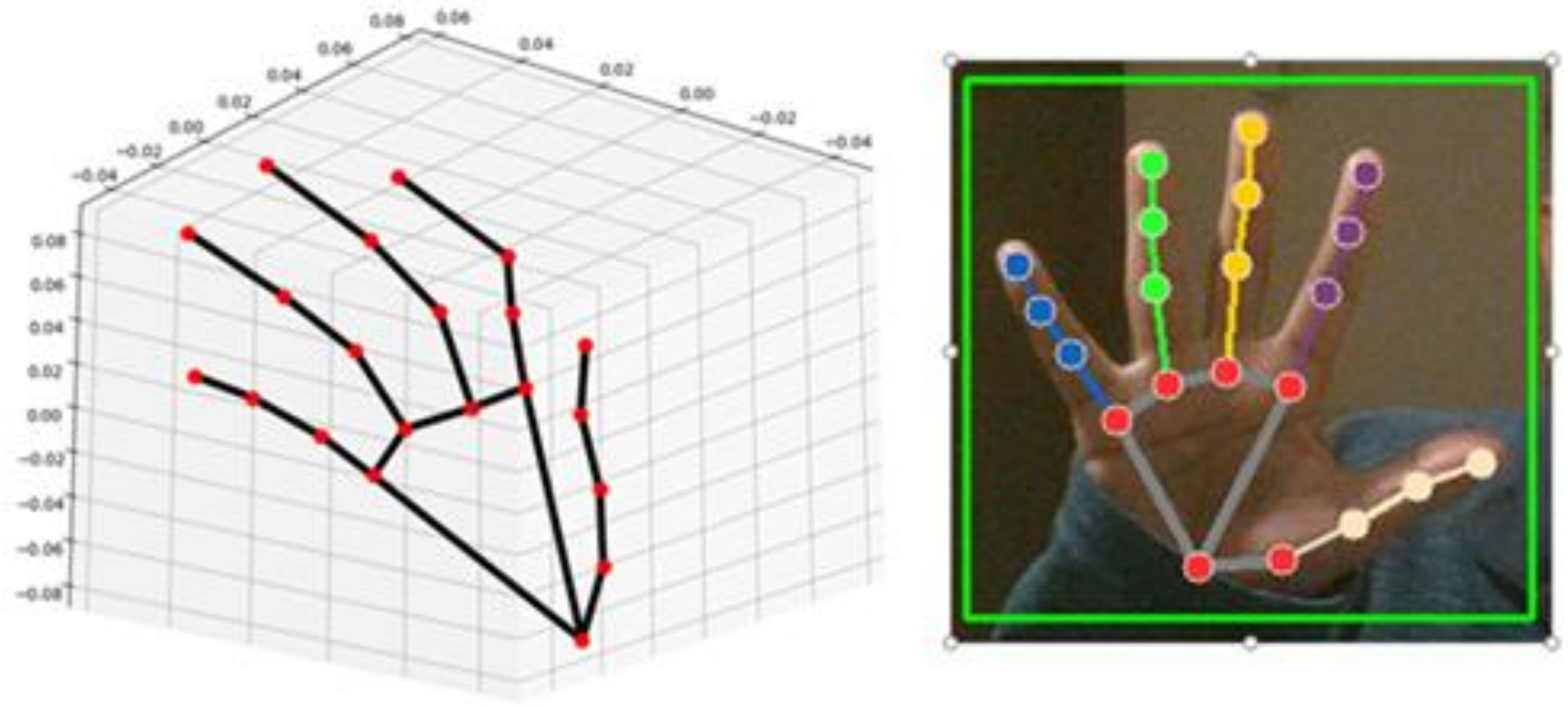

Fig. 5 Hand pose estimation using Google Mediapipe

The hand pose also gives landmarks and these are basically 3D landmarks with X and Y coordinates as per the video frame size, but the Z value is a scaled value i.e., not a true depth. The wrist is an anchor in this pose estimation meaning the coordinates of the wrist landmark is (X, Y, 0) i.e., the wrist is always at depth equals to 0 i.e., Z = 0. All the rest of the landmarks are referenced off the wrist Z value. Several techniques were thought and tested for calculating the angle of the finger flexion and extension. The first approach was finding the Euclidean distance between the fingertips and MCP and relate it to scaling of the angle the finger motor can move during actuation. This failed because with distance to camera, the length would change. Same issue was caused when cosine law was used. Vector dot product was used to find this angle

$$\theta = \cos^{-1}\left[\frac{(a.b)}{||a||||b||}\right] \tag{4}$$

Where θ is the angle, cos-1 is the inverse cosine, a and b are two vectors ||a|| and ||b|| are normalized vectors. For (4) to work, the coordinates for fingertip and MCP and Wrist and MCP are converted to vectors and fed in to (4). The wrist rotation was a challenge because the axis of rotation was the X-axis. The gradient of wrist to index MCP and thumb MCP are calculated and fed into the formula as below keeping sure that the denominator never goes to zero

$$\theta = \tan^{-1}\left[\frac{(m_1 - m_2)}{(1 + m_1 \times m_2)}\right] \tag{5}$$

Where θ is the angle, tan-1 is the inverse tan, m1 and m2 are the gradients. If the denominator goes to zero, the program converts the denominator to 1 thus not giving a Divide by Zero exception.

These values are then fed into an averaging filter to smooth the values of the angles calculated as in (6), where N is the number of values taken, for the filtration N = 3 was used, ang_val is the value of a joint's angle.

$$Average\ angle = \frac{1}{N!} \times \sum_{i=1}^{N} i \times ang_val\,[\ :i+1\,] \tag{6}$$

These averaged values were then mapped by using (7) where it takes in 5 values, the actual input as x, then there are 4 ranges of input and output as below

$$Coverted\ Value = max\begin{cases} min\begin{cases} out_max \\ \left\lfloor \frac{out_{max}-out_{min}}{in_{max}-in_{min}} \right\rfloor + out_min \end{cases} \\ out_min \end{cases} \tag{7}$$

These values are sent to the Arduino through PyFirmata, and the fingers actuate as per these values.

# 3. RESULTS

The result discusses three sections describing the results of the respective methodologies.

### 3.1. Electromechanical Build

Since the mechanical build has deviated a lot from the original design. The major changes were the servo pulley from single track to dual track pulley as in figure 6, and the tautness of the tendons was increased. The mass of all pulleys was 10 grams.

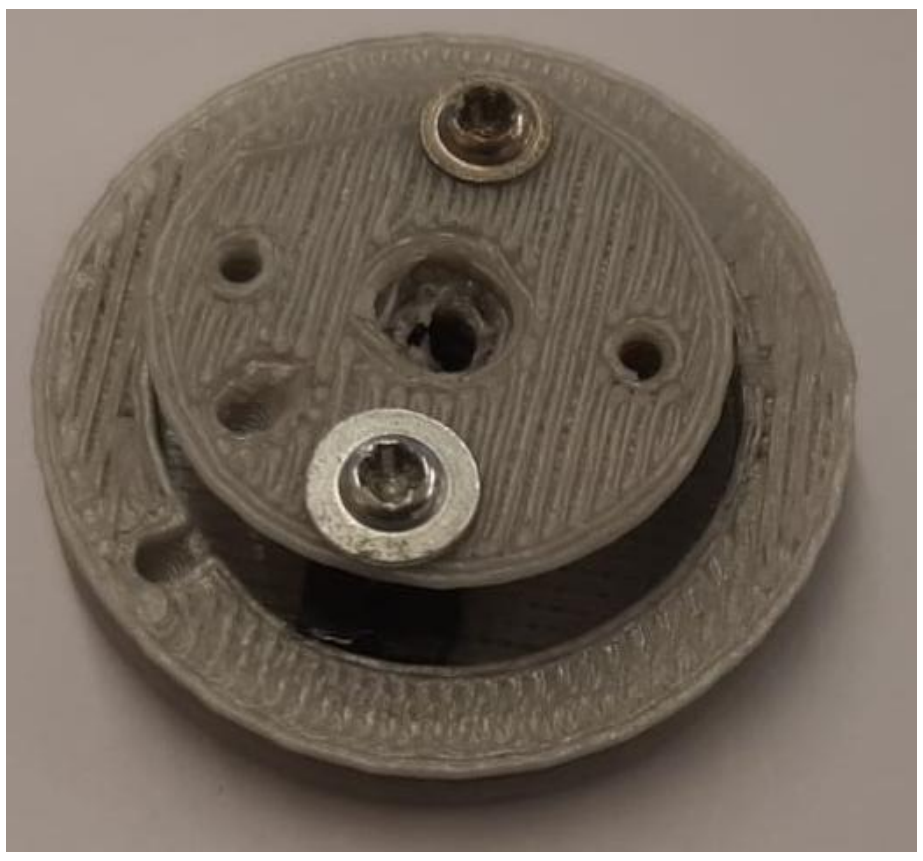

Fig. 6 New pulley design with dual tracks

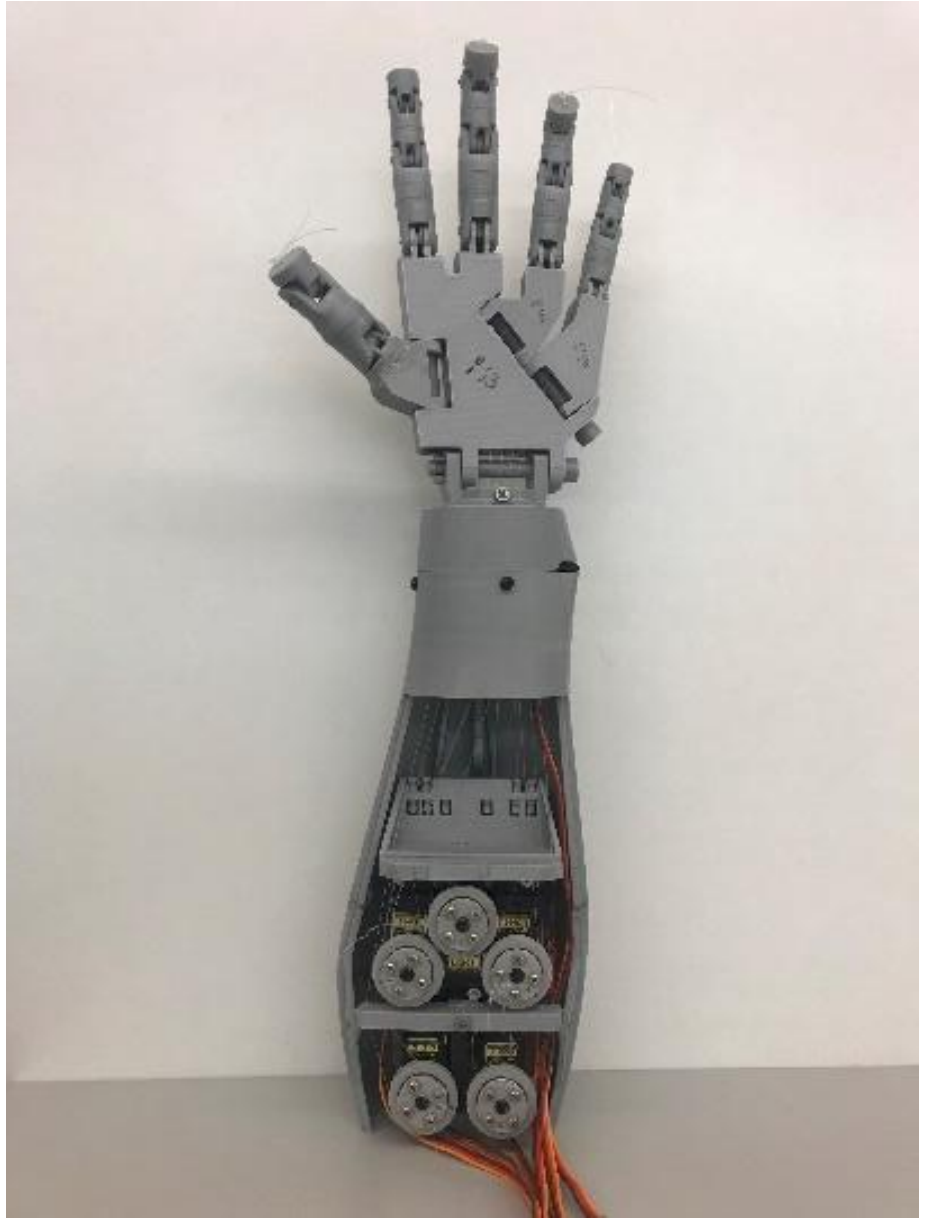

Fig. 7 Forearm with installed motor and tendons

The bicep motor plate was swapped for other servo motor and required another motor mounting plate, below is the new motor plate as in figure 8 design and it has a mass of 20 grams.

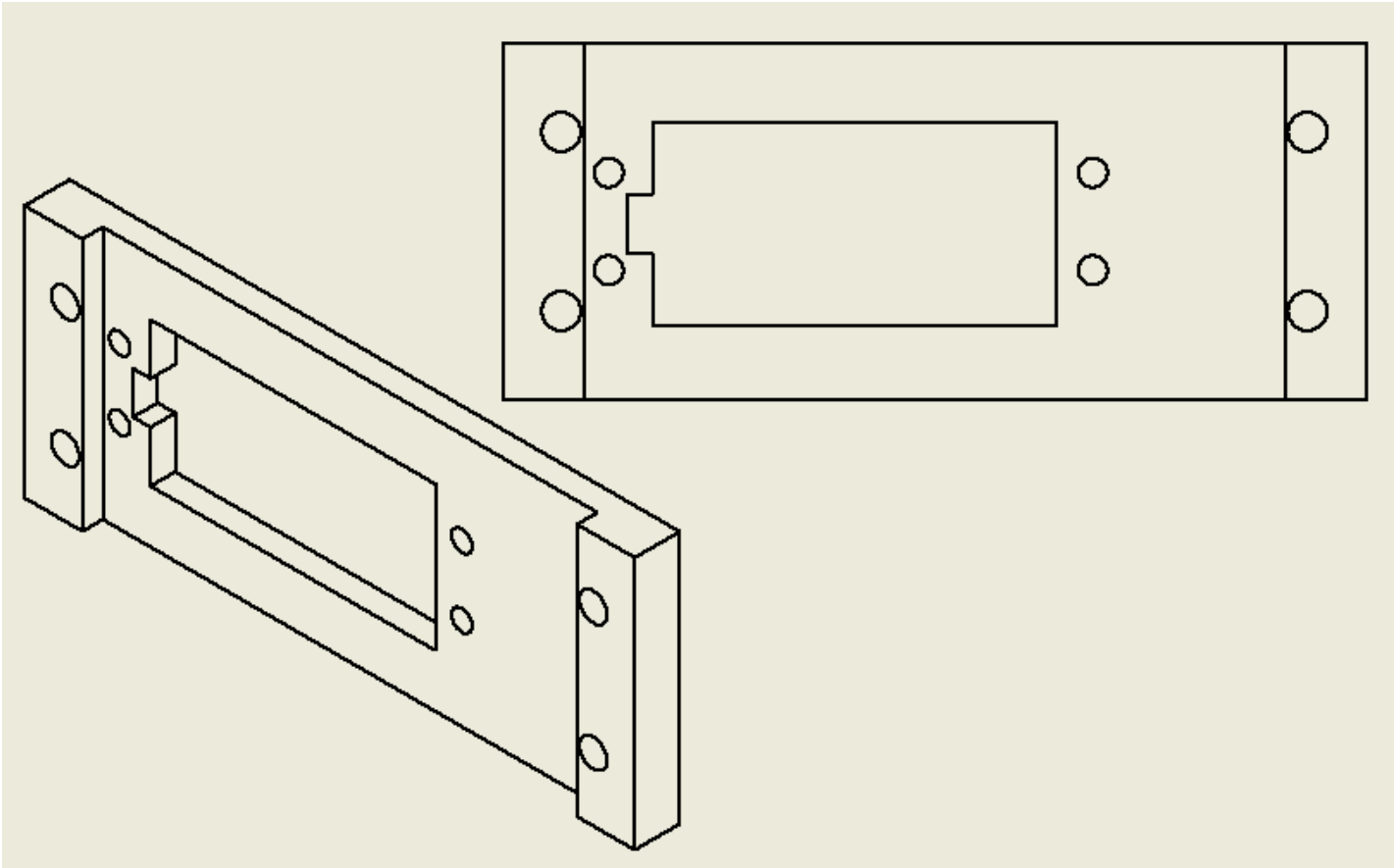

Fig. 8 Bicep motor plate

The elbow motor was also swapped so a new plate was designed as in figure 9. It has a mass of 20 grams.

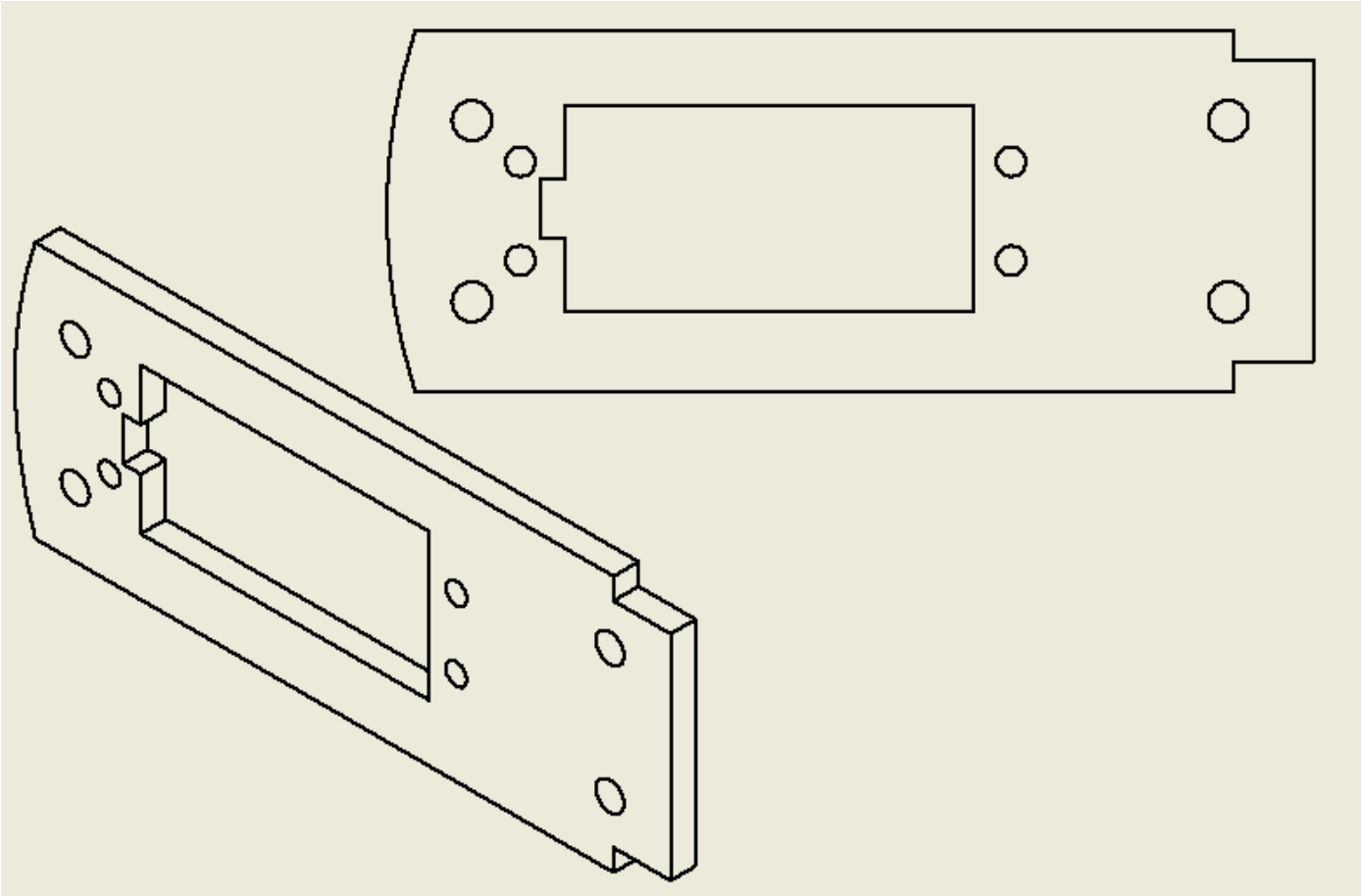

Fig. 9 Elbow motor plate

The original motors as per the Inmoov design was Hitec HS-805B and has a mass of 152 grams. So, a total of 304 grams was reduced off the arm and 40 grams was added in its place. Since the new bracket is not part of the arm, the mass did not make any effect on the operation of the arm. The two new motors added are TD8125 and has a combined mass of 112 grams. The new motors have no difference in the performance and work greatly. The new motor provides 23.5 kg/cm of torque at 4.8 V which more compared to 19.8 kg/cm.

The one issue that arises from this upgrade is that the original parts need to be modified for the new one because the body of the motor provides rigidity to the motor holder which right now is not there and can be removed by either by redesigning (making the motor slot smaller) the part or filling the gaps in the original part. However, there is next to no vibration caused by it and works smoothly. There is no observable buckling or vibrations in the structure. The total mass of the arm came out to be 2168 grams. This does not include the bracket designed for mounting the arm otherwise the mass is around 2700 grams.

### 3.2. Mechanism and Kinematics

The results for this section entail the results achieved by simulation and the software developed and comparison of the simulation posture and the real-world robot. The figure 10 below shows the robot's work volume

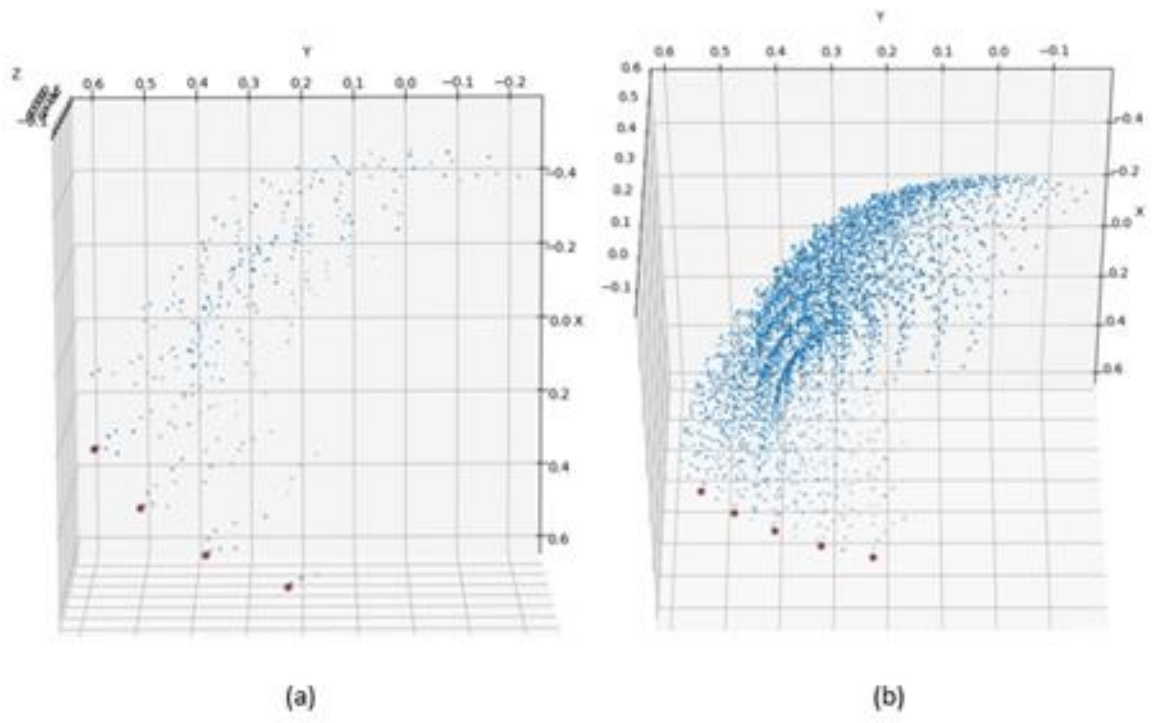

Fig. 10 Work volume of the arm

Following is the result of the software developed and the posture. From Table 4, Motion 1 and 5 are shown in the figures below.

TABLE 4. DH Parameters for the arm

| | Joint Variables | | | | | MRPT | | | Python | | |
|---|---|---|---|---|---|---|---|---|---|---|---|
| | θ1 | θ2 | θ3 | θ4 | θ5 | X | Y | Z | X | Y | Z |
| 1 | 0 | -90 | -90 | 26.6 | 0 | 0.567 | 0.232 | 0 | 0.567 | 0.232 | 0 |
| 2 | 20 | -90 | -90 | 26.6 | 0 | 0.454 | 0.412 | 0 | 0.454 | 0.412 | 0 |
| 3 | 0 | -135 | -90 | 26.6 | 0 | 0.405 | 0.232 | 0.393 | 0.405 | 0.232 | 0.393 |
| 4 | 0 | -135 | -40 | 26.6 | 0 | 0.315 | 0.173 | 0.482 | 0.315 | 0.173 | 0.482 |
| 5 | 35 | -215 | -40 | 66.6 | 0 | -0.503 | -0.005 | 0.155 | -0.503 | -0.005 | 0.155 |

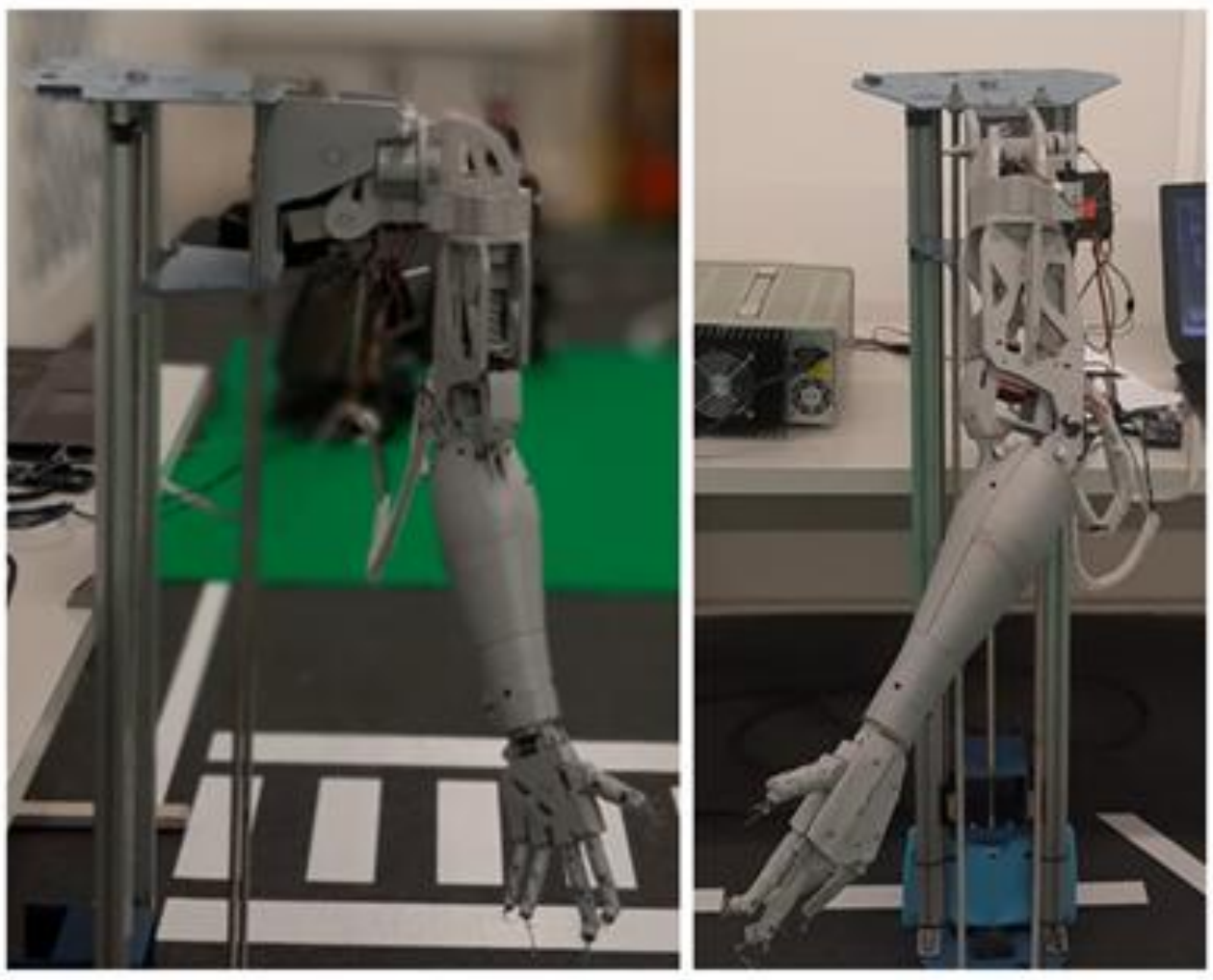

Fig. 11 Arm configuration as of 1 from Table 4

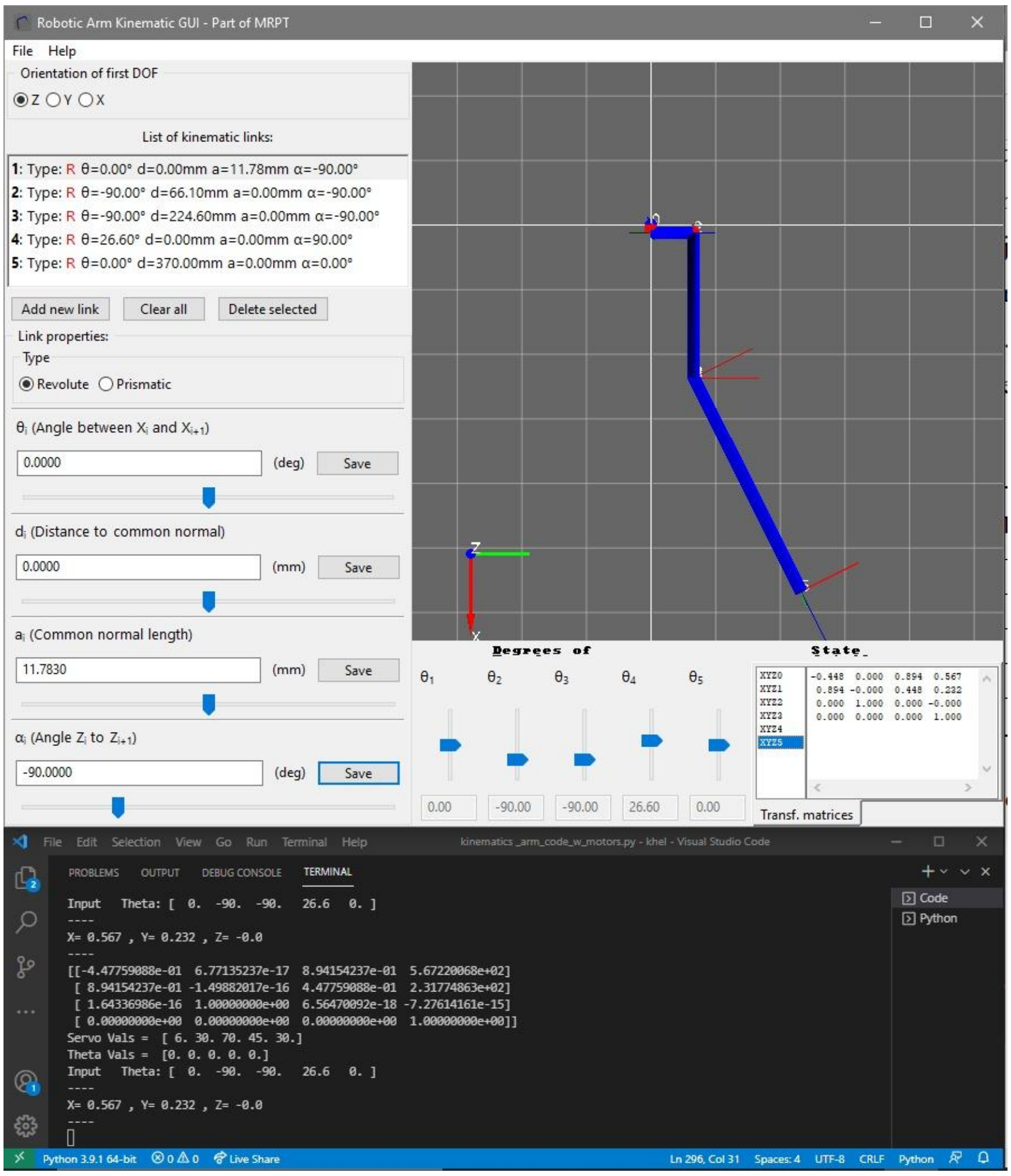

Fig. 12 Comparison of software and simulation

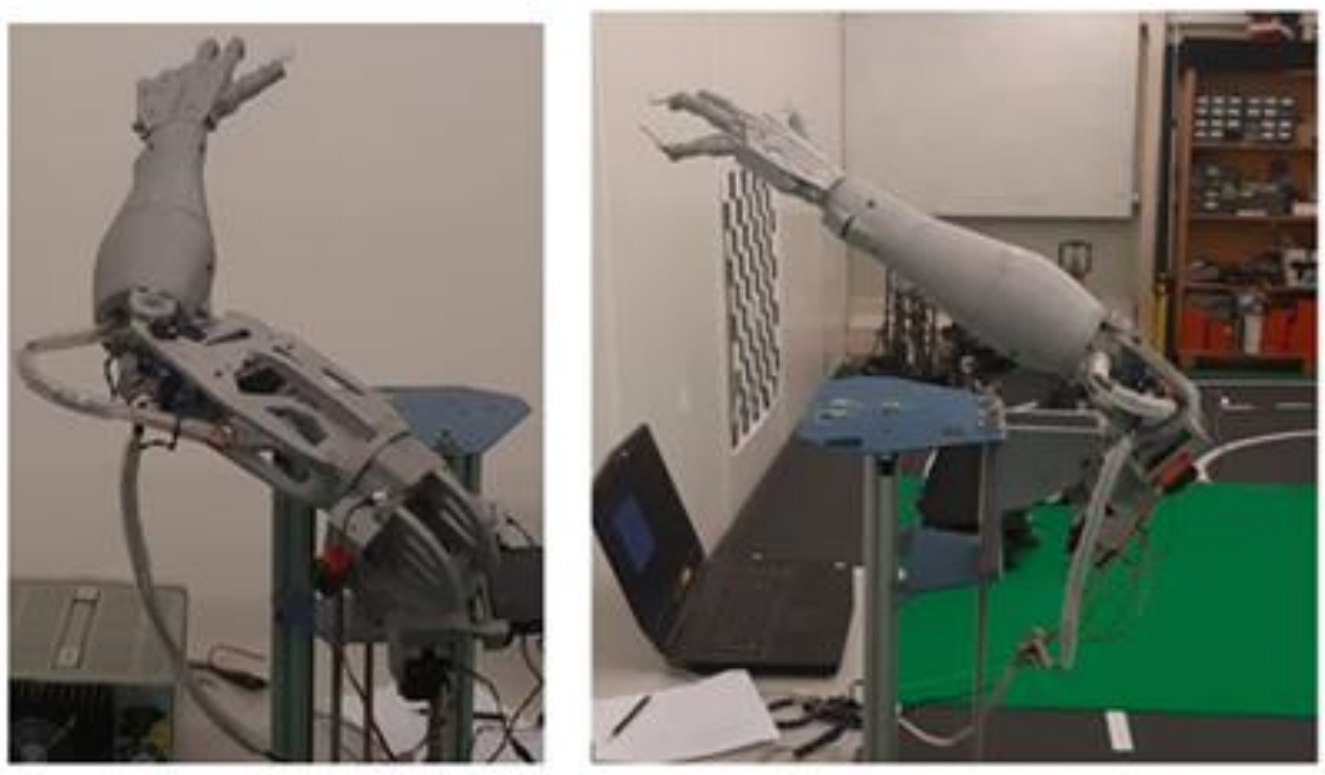

Fig. 13 Arm configuration as of 5 from Table 4

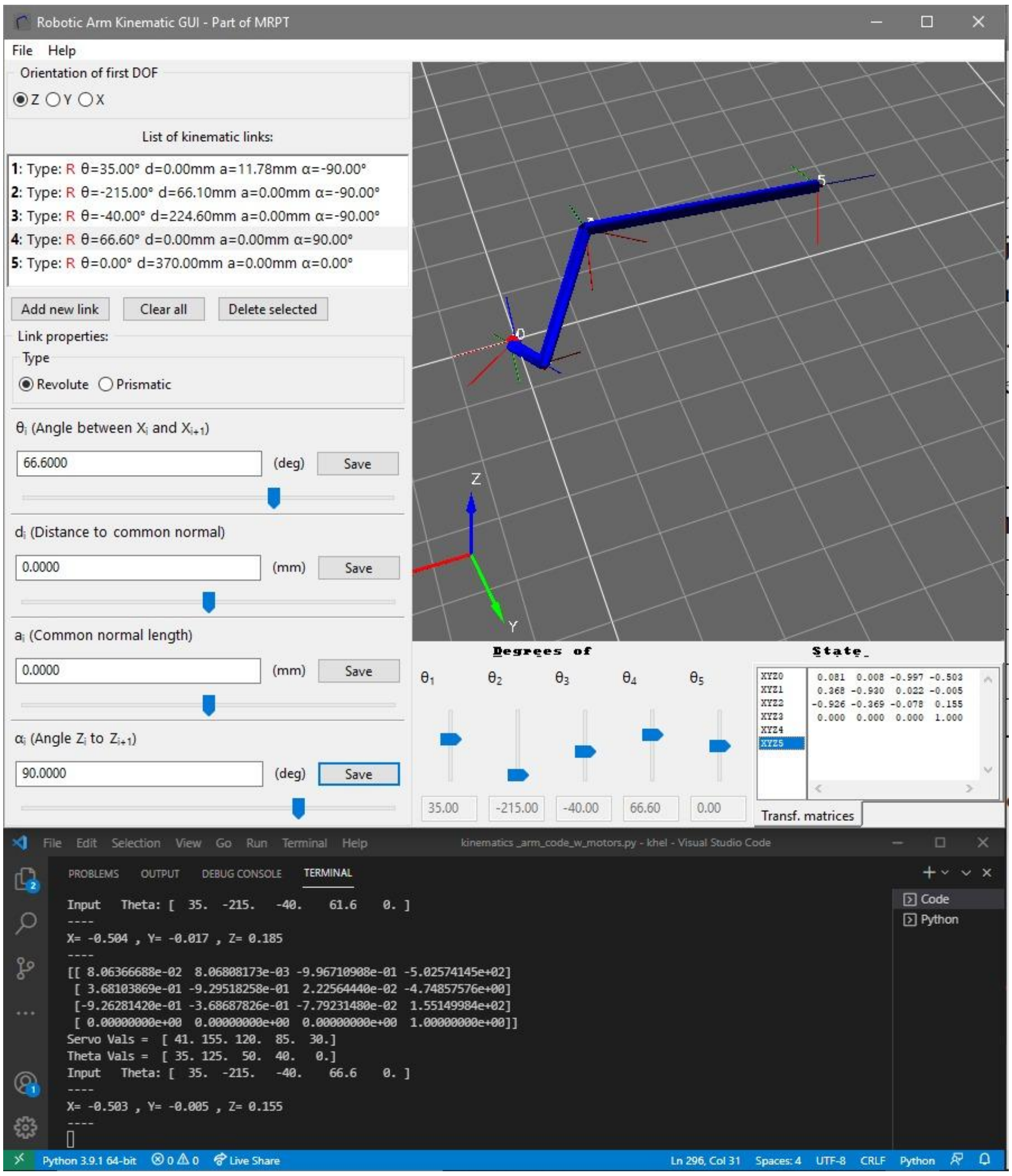

Fig. 14 Comparison of software and simulation

(b)

### 3.3. Mimicking through Vision

The computer vision aspect of the work details the teleoperation capability of the robot. The practical work gave good fps ranging from 16 to 25. The range of the motion of servos is discussed in the methodology and that range applies to complete motion. The motion of the hand as per the camera input is shown below. The figure 16 shows two things, firstly, the five fingers of the hand and the hand position and the wrist movement. Wrist movement is not all 180-degree actuation with respect to the servo motor, rather a 120 degree of motion from 30 to 150 degree is allowed. The reason being is the load on the tendons go through the wrist and get taut and loose based on the wrist position.

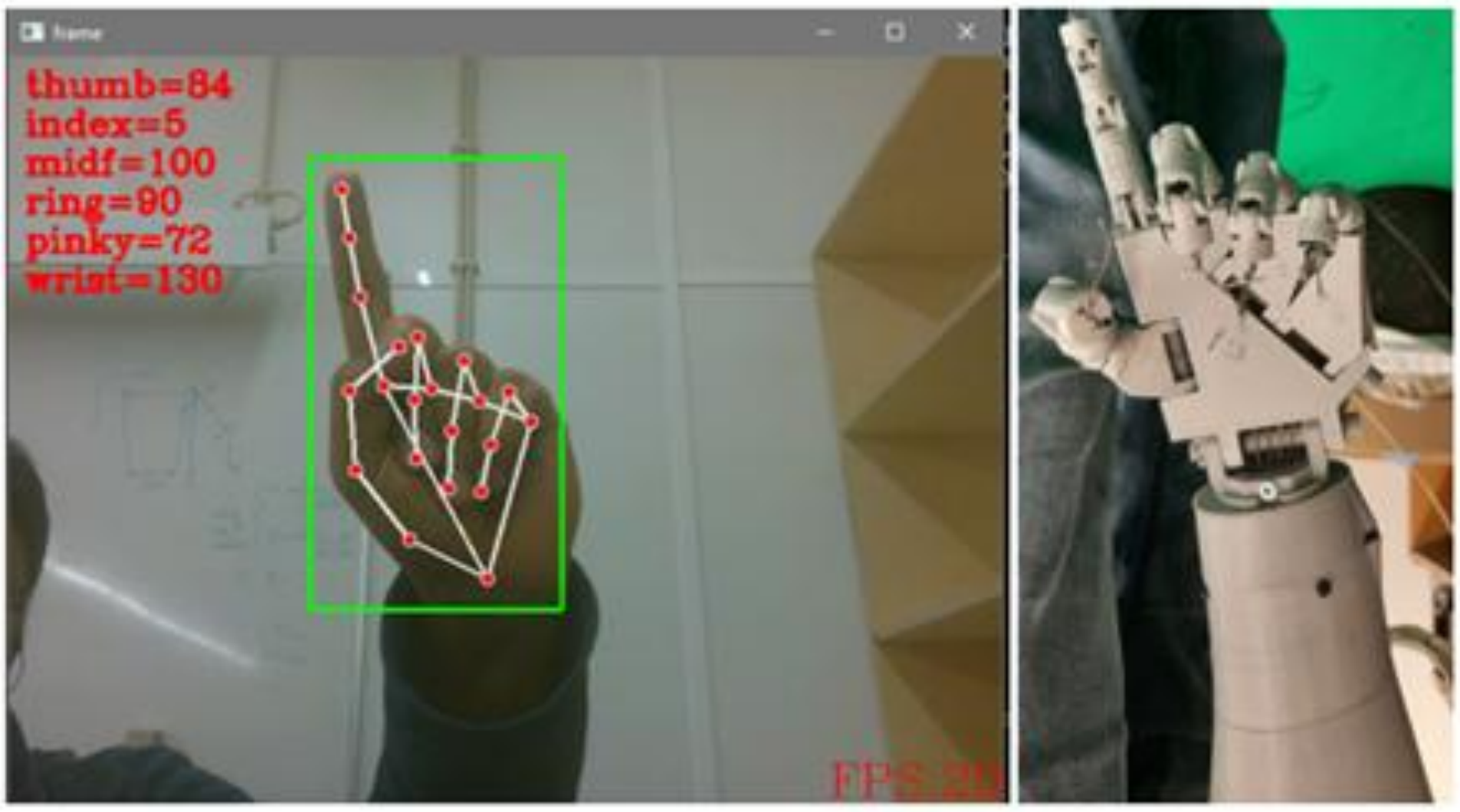

(a)

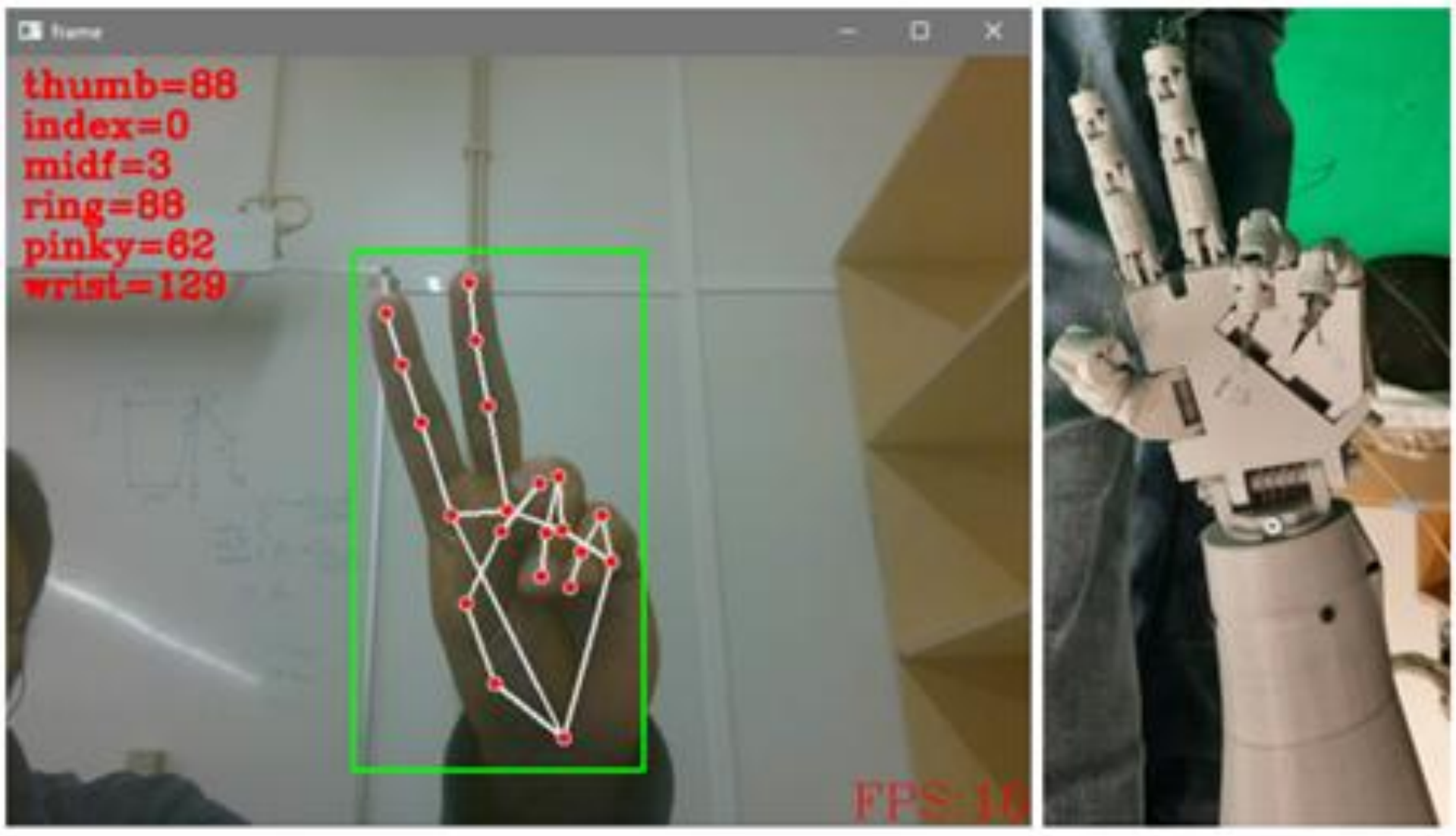

(b)

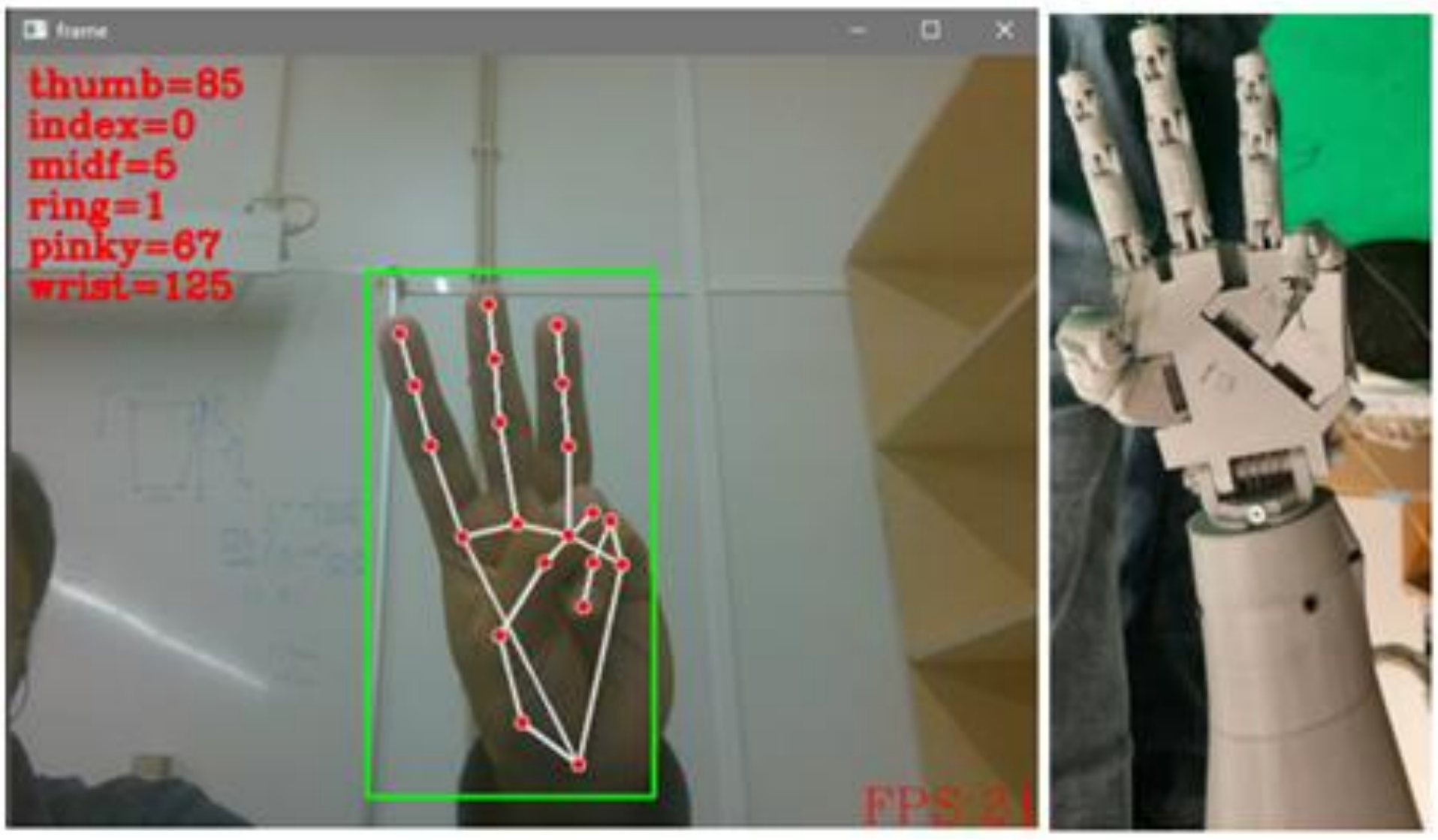

(c)

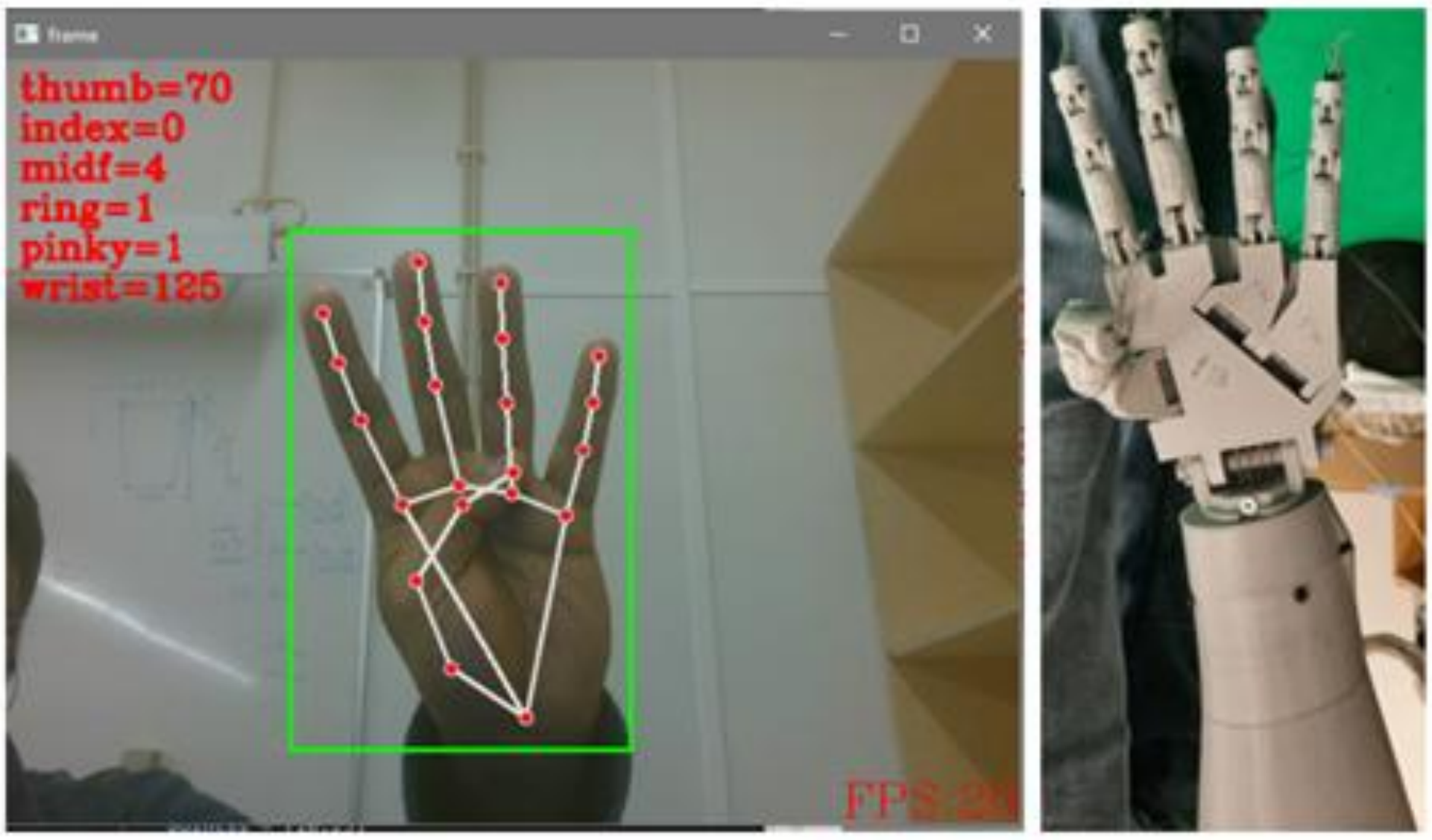

(d)

Fig. 15 (a - d) Holding up fingers and robot mimicking.

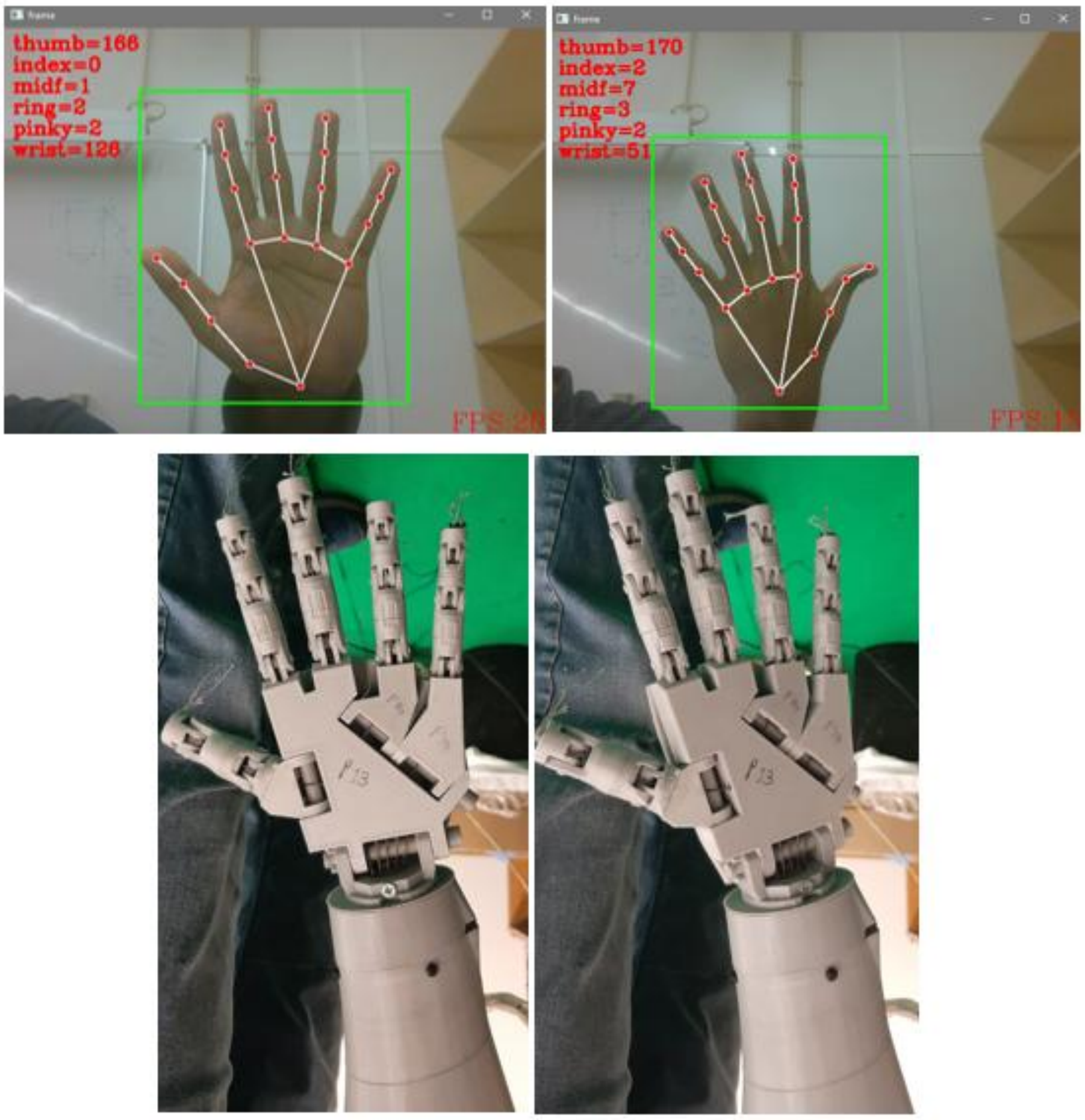

Fig. 16 Wrist rotation and the position of the hand.

# 4. CONCLUSIONS

The paper discusses the development of a robotic arm and the development of the kinematics for the Inmoov arm being utilized in CHARMIE. The paper also entails the concept of teleoperation via computer vision. The future development for this work is to utilize a 3D vision system to develop the mimicking aspect and the teleoperation of the robot and to implement the inverse kinematic for the robot. The two works can be merged i.e., the 3D computer vision and the inverse kinematics to enable the robot to interact with the real word scenarios.

# 5. FURTHER WORK

This work is the basis for the development of the CHARMIE robot and it establishes a starting point for the development of arms and hand and their control in the future iteration, It introduced several avenues for control strategies such as fully automated that is the system uses predefined instruction and action that can operate in the real world, also a method for teleoperation of the robot, where a human counterpart can control the robot either online or throw a pendant. The upgrades that have been naturally selected and are planned to be studied in the future are as follows:

- The need for improvement in the robot arm actuation i.e. the lead screw mechanism in the elbow and the shoulder actuation to a machine screw mechanism which will provide more friction to the holding of the position and relieving the motor after an actuation.
- A better servo motor with the improved sensor for measurement such as an encoder or very accurate potentiometer for position control.
- Use of 3D camera instead of 2D camera for better depth measurement as it give one more dimension for control of the hand pose estimation and further it can be implemented to the whole arm and body by using body pose estimation.
- Development of inverse kinematic for implementation of automated control of the arm and development of parallel forward and inverse kinematics of the robot hand.
- Improvement of the mechanical systems with respect to strength, material, and design aspects.

# 6. ACKNOWLEDGEMENTS

The authors would like to acknowledge the Laboratory of Automation and Robotics (University of Minho, Guimarães) for their support, equipment, and valuable staff assistance.

"This work has been supported by FCT – Fundação para a Ciência e Tecnologia within the R&D Unit Project Scope UID/00319/2025 - Centro ALGORITMI (ALGORITMI/UM)", and the R&D Units Project Scope: UIDB/00319/2020, by a doctoral scholarship from the Portuguese Foundation for Science and Technology (Fundação para a Ciência e a Tecnologia) [grant number SFRH/BD/06944/2020], with funds from the Portuguese Ministry of Science, Technology and Higher Education and the European Social Fund through the Programa Operacional do Capital Humano (POCH).